\documentclass[10pt,twocolumn,letterpaper]{article}

\usepackage{wacv}      
\usepackage{multirow}
\usepackage{amssymb}
\definecolor{wacvblue}{rgb}{0.21,0.49,0.74}
\usepackage[pagebackref,breaklinks,colorlinks,allcolors=wacvblue]{hyperref}

\def\wacvPaperID{*****} 
\def\confName{WACV}
\def\confYear{2026}

\title{nnMobileNet++: Towards Efficient Hybrid Networks for Retinal Image Analysis }

\author{
Xin Li\\
Arizona State University\\
{\tt\small x.li@asu.edu}
\and
Wenhui Zhu\\
Arizona State University\\
{\tt\small wzhu59@asu.edu}
\and
Xuanzhao Dong\\
Arizona State University\\
{\tt\small xdong64@asu.edu}
\and
Hao Wang\\
Clemson University\\
{\tt\small hao9@g.clemson.edu}
\and
Yujian Xiong\\
Arizona State University\\
{\tt\small yxiong42@asu.edu}
\and
Oana M. Dumitrascu\\
Mayo Clinic\\
{\tt\small Dumitrascu.Oana@mayo.edu}
\and
Yalin Wang\\
Arizona State University\\
{\tt\small ylwang@asu.edu}
}

\begin{document}
\maketitle
\begin{abstract}
Retinal imaging is a critical, non-invasive modality for the early detection and monitoring of ocular and systemic diseases. Deep learning, particularly convolutional neural networks (CNNs), has significant progress in automated retinal analysis, supporting tasks such as fundus image classification, lesion detection, and vessel segmentation. As a representative lightweight network, nnMobileNet has demonstrated strong performance across multiple retinal benchmarks while remaining computationally efficient. However, purely convolutional architectures inherently struggle to capture long-range dependencies and model the irregular lesions and elongated vascular patterns that characterize on retinal images, despite the critical importance of vascular features for reliable clinical diagnosis. To further advance this line of work and extend the original vision of nnMobileNet, we propose nnMobileNet++, a hybrid architecture that progressively bridges convolutional and transformer representations. The framework integrates three key components: (i) dynamic snake convolution for boundary-aware feature extraction, (ii) stage-specific transformer blocks introduced after the second down-sampling stage for global context modeling, and (iii) retinal image pretraining to improve generalization. Experiments on multiple public retinal datasets for classification, together with ablation studies, demonstrate that nnMobileNet++ achieves state-of-the-art or highly competitive accuracy while maintaining low computational cost, underscoring its potential as a lightweight yet effective framework for retinal image analysis.

\end{abstract}

\begin{figure}[h]
  \centering
\centerline{\includegraphics[width=0.5\textwidth]{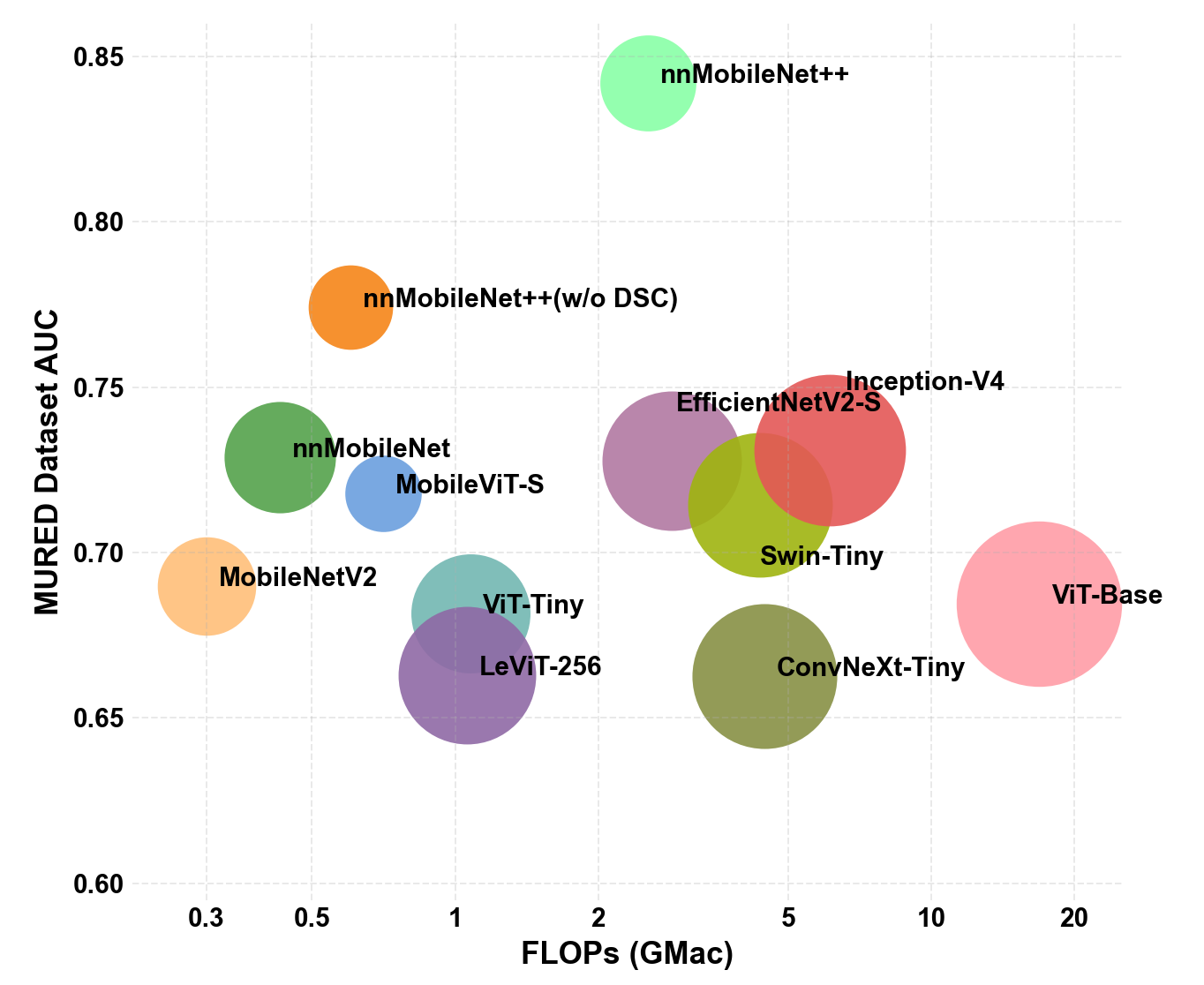}}
\caption{
Comparison of FLOPs, parameter size, and AUC on the MURED dataset. 
Each bubble corresponds to a model, where the \textbf{bubble size} is proportional to the number of parameters, 
the \textbf{x-axis} indicates computational cost (FLOPs), and the y-axis shows classification performance (AUC). The nnMobileNet++ (w/o DSC) denotes the architecture after removing the Dynamic Snake Convolution module.
}
\label{fig:flops}
\end{figure}
    
\section{Introduction}
\label{sec:intro}

Retinal imaging plays a critical role in modern medicine as a non-invasive and accessible modality for the early detection and longitudinal monitoring of ocular and systemic diseases~\cite{wolf2020cost,wang2025artificial,dong2025tpot,zhu2023otre,zhu2023optimal}. Retinal disorders such as diabetic retinopathy, glaucoma, and age-related macular degeneration remain among the leading causes of visual impairment and blindness worldwide, underscoring the urgent need for reliable diagnostic tools~\cite{world2006prevention,steinmetz2021causes,congdon2004causes,vasa2025context,dong2025cunsb, wang2024many,dong2024dme}. Beyond ophthalmology, retinal images also provide important biomarkers for systemic conditions including hypertension, diabetes, and neurodegenerative diseases such as Alzheimer’s disease, highlighting their broad clinical relevance~\cite{DumitrascuAD, dumitrascu2024color,dumitrascu2018retinal}. In recent years, deep learning–based approaches have become the dominant paradigm for automated retinal image analysis, with convolution neural networks (CNNs) supporting tasks such as disease classification, lesion detection, and vessel segmentation~\cite{zhu2024nnmobilenet,dong2024dme,wang2024many,li2025evit,dong2025tpot,dumitrascu2024color,zhu2024selfreg,dong2025cunsb,zhu2025retinalgpt,zhu2024selfreg}. CNNs are effective for retinal image analysis using convolution operations extract features at the pixel level, enabling efficient detection of retinal structures such as lesions and vessels, while keeping the models sufficiently efficient for deployment on resource-constrained or edge devices~\cite{zhu2024nnmobilenet, li2025evit, sandler2018mobilenetv2,tan2019efficientnet}. Among these efforts, nnMobileNet has emerged as a representative lightweight baseline, delivering competitive accuracy with low parameter counts and computational cost through a careful redesign of CNN components~\cite{zhu2024nnmobilenet}. However, purely convolution architectures have limited ability to model long-range dependencies, which are essential for capturing the irregular lesions and spatially distributed vascular patterns commonly observed in retinal images.~\cite{dosovitskiy2020image,liu2021swin,mehta2021mobilevit}

Vision Transformers (ViTs) have recently demonstrated notable capability in global context modeling across a range of computer vision benchmarks, effectively capturing long-range dependencies. In medical imaging, ViTs have been applied to tasks such as segmentation, organ delineation, and disease classification, showing steady progress~\cite{dosovitskiy2020image,liu2021swin,mehta2021mobilevit,li2025evit,graham2021levit,cao2022swin,zhu2024selfreg}. For retinal analysis, global reasoning is particularly critical: Lesions often have irregular shapes and scattered spatial distribution, while vessels are thin and tortuous, and vascular characteristics are closely associated with ocular diseases and systemic conditions such as Alzheimer’s disease. These heterogeneous features require models to preserve fine boundary information while maintaining reliable global reasoning~\cite{lim2017retinal,fathimah2025retinal,liu2025association,dumitrascu2018retinal,dumitrascu2024color,frost2013retinal,zhou2025early,wang2024octa,chua2019impact}.
Nonetheless, routine application of ViTs in medical imaging remains constrained. Stable convergence often depends on large-scale, high-quality annotations, which are scarce and costly in clinical practice; moreover, the quadratic complexity of self-attention results in considerable computational and memory overhead, limiting deployment on resource-constrained diagnostic systems~\cite{dosovitskiy2020image,liu2021swin,zhu2024nnmobilenet,li2025evit}. These limitations motivate hybrid frameworks that retain the lightweight efficiency of CNNs while incorporating the contextual modeling strengths of ViTs. To further enhance vascular representation, we introduce Dynamic Snake Convolution (DSC)~\cite{qi2023dynamic}, enabling adaptive sampling along curvilinear structures. This helps preserve vessel continuity and morphology, and provides structured features more effective in subsequent ViTs.

Motivated by the need to combine the efficiency of CNNs with the global dependency modeling of ViTs, we extend nnMobileNet~\cite{zhu2024nnmobilenet} and propose nnMobileNet++, a hybrid framework that integrates local convolution representations with transformer-based global reasoning. This work makes the following key contributions: (i) an efficient hybrid architecture that integrates convolution and transformer components while maintaining lightweight design, (ii) the introduction of dynamic snake convolution~\cite{qi2023dynamic} together with Vision Transformer modules before the Vision Transformer stage, which enhances boundary-sensitive local feature extraction and preserves curvilinear structures such as blood vessels during downsampling, and (iii) self-supervised pretraining on a large-scale retinal image dataset to enhance robustness and generalization.
We evaluated the framework on multiple retinal image datasets, and the results demonstrate a favorable balance between accuracy and efficiency (Fig.~\ref{fig:flops}), highlighting its practicality for retinal image analysis in clinical settings.

\section{Related work}
\label{sec:related_work}
\subsection{Lightweight CNN-Based Models}
Lightweight CNNs have been widely adopted in retinal image analysis due to their efficiency and suitability for deployment on fundus cameras and edge devices~\cite{li2025evit, zhu2024nnmobilenet, sandler2018mobilenetv2, tan2019efficientnet,liu2022convnet}. Building on earlier models such as MobileNet~\cite{sandler2018mobilenetv2} and EfficientNet~\cite{tan2019efficientnet}, nnMobileNet~\cite{zhu2024nnmobilenet} introduces task-specific refinements, including stage-wise channel allocation~\cite{han2021rethinking}, heavy data augmentation strategies, AdamP optimization~\cite{heo2020adamp}, spatial dropout, and ReLU6 activation~\cite{zhu2024nnmobilenet}.These improvements enhance both accuracy and robustness across retinal benchmarks, while keeping computational demands low, establishing nnMobileNet as an efficient baseline. However, as a purely convolution model, nnMobileNet remains constrained in modeling long-range dependencies, which are essential for capturing complex retinal structures~\cite{dosovitskiy2020image}.

\subsection{Transformer-Based and Hybrid Models}
Vision Transformers (ViTs) were originally proposed as an alternative to convolution architectures, introducing self-attention mechanisms that enable explicit modeling of long-range dependencies and global context~\cite{dosovitskiy2020image}. These strengths quickly led to widespread exploration in medical imaging, where ViTs have shown promising results in tasks such as organ segmentation, tumor delineation, and disease classification~\cite{chen2021transunet,hatamizadeh2022unetr,wang2021transbts}. Nevertheless, standard ViT architectures require large annotated datasets to converge reliably and impose high computational and memory demands, which limit their adoption in clinical practice and on resource-constrained devices.

Hybrid architectures such as MobileViT~\cite{mehta2021mobilevit} and LeViT~\cite{graham2021levit} in the vision community have shown that combining convolution with lightweight attention can provide a good balance between accuracy and efficiency. While related ideas have also been explored in medical imaging, most existing designs were developed for generic benchmarks and lack adaptations to the structural characteristics of retinal images~\cite{li2025evit, cao2022swin}. Given nnMobileNet’s strong performance and lightweight design on retinal images~\cite{zhu2024nnmobilenet}, we adopt it as our baseline and incorporate Vision Transformer modules to enhance global dependency modeling.

\begin{figure*}[h]
  \centering
  \centerline{\includegraphics[width=0.95\textwidth]{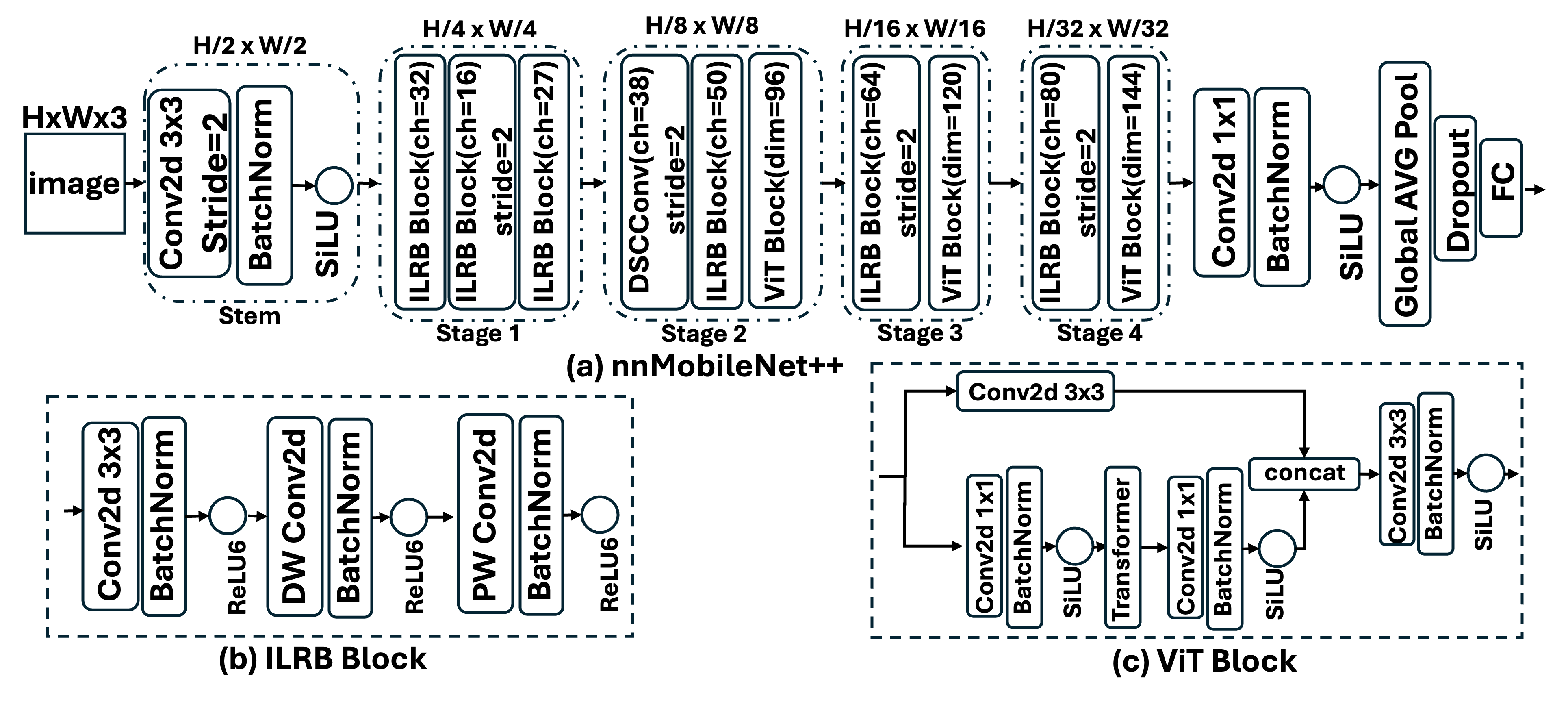}}
\caption{
(a) Overall architecture of our network, consisting of four stages where early stages are CNN-based and later stages are ViT-based. At the second-stage downsampling, a \textbf{Dynamic Snake Convolution (DSC)}~\cite{qi2023dynamic} is employed to better preserve curvilinear structures before feeding features into the ViTs. (b) The CNN blocks are built from nnMobileNet’s Inverted Residual Linear Bottleneck (IRLB), where \textbf{Depthwise (DW)} convolutions capture spatial information and \textbf{Pointwise (PW)} convolutions perform channel mixing. (c) The ViT modules combine local convolutions with global self-attention for contextual representation. ``\textbf{ch}'' denotes the number of input channels.
}
\label{fig:arch}
\end{figure*}

\subsection{Retinal-Specific Architectures}
Retinal images present irregular lesions and thin, tortuous vessels that span large regions, with vascular morphology serving as a recognized biomarker for ocular and systemic diseases~\cite{lim2017retinal,fathimah2025retinal,liu2025association,dumitrascu2018retinal,frost2013retinal,zhou2025early,wang2024octa,chua2019impact,wang2025retinal,wang2024rbad}. While convolution neural networks (CNNs) are effective for local feature extraction, their limited receptive fields and downsampling can distort fine vascular structures. Vision Transformers (ViTs) naturally capture long-range dependencies and global context, making them well suited for distributed retinal features~\cite{luo2016understanding,liu2021swin,dosovitskiy2020image}. To preserve structural detail while retaining CNN efficiency, we incorporate dynamic snake convolution(DSC)~\cite{qi2023dynamic} in the convolution stage. Standard convolutions use fixed kernels and often oversmooth or distort curved boundaries, whereas DSC is boundary-sensitive and adapts to curvilinear patterns~\cite{qi2023dynamic}. This makes it particularly effective for modeling retinal vessels and irregular lesions, ensuring that such structures are preserved and passed to the ViT stage, where global dependencies can then be captured on top of reliable local features.

\subsection{Self-Supervised Pretraining in Retinal Imaging}
Medical imaging datasets are typically small, and labeled retinal data are especially limited due to the cost of expert annotation~\cite{zhu2024nnmobilenet,li2025bert, hervella2020learning}. Self-supervised pretraining has emerged as an effective way to address this limitation and has shown benefits across modalities such as MRI, CT, and fundus photography~\cite{azizi2021big,dumitrascu2024color,li2025bert,tang2022self}. Within this line of work, generative masked image modeling has proven particularly suitable for medical imaging, as the reconstruction of missing regions encourages the network to capture both fine local details and broader structural patterns that are central to anatomical representation~\cite{he2022masked,xie2022simmim,gupta2024medmae}. Building on this insight, we adopt SimMIM~\cite{xie2022simmim} for pretraining on large collections of unlabeled retinal data. Its design places no restrictions on network structure, making it well suited to our hybrid CNN–ViT framework and enabling the model to learn structural priors at multiple scales, thereby improving robustness and generalization across datasets.

\section{Method}
\label{sec:method}

\subsection{Overall Architecture}
As illustrated in Fig.~\ref{fig:arch}, we introduce nnMobileNet++, a four-stage hybrid network. The network begins with a stem convolution for low-level feature extraction, followed by four sequential stages. Stage 1 and Stage 2 retain the inverted residual linear bottleneck (IRLB) design of nnMobileNet~\cite{zhu2024nnmobilenet}, which use depthwise convolutions for spatial feature extraction and pointwise convolutions for channel mixing,  achieving a  trade-off between accuracy and efficiency~\cite{zhu2024nnmobilenet,han2021rethinking}. At the downsampling step of Stage 2, we introduce a Dynamic Snake Convolution (DSC)~\cite{qi2023dynamic} to better preserve curvilinear retinal structures such as blood vessels before passing features into the transformer modules. Stage 3 and Stage 4 are built with Vision Transformer modules~\cite{dosovitskiy2020image,mehta2021mobilevit}, combining local convolutions with global self-attention to capture both fine-grained spatial details and long-range dependencies. Across the four stages, spatial resolution is  reduced via IRLB-based downsampling, leading to a final downsampling factor of 32 relative to the input. Finally, a lightweight classification head aggregates hierarchical features for prediction. Importantly, nnMobileNet++ preserves the non-traditional but effective channel configuration~\cite{han2021rethinking} of nnMobileNet, as shown in Fig.~\ref{fig:arch}(a). Overall, this design achieves a strong balance between efficiency and representational power, while retinal-specific pretraining further improves downstream task performance.

\subsection{Dynamic Snake Convolution (DSC) for Vessel Structures}
Standard convolution focuses on local neighborhoods, which can break vessel continuity—especially during downsampling~\cite{ronneberger2015u,dai2017deformable,qi2023dynamic}. To mitigate this, we adopt Dynamic Snake Convolution (DSC)~\cite{qi2023dynamic}, which adaptively models sampling offsets along curvilinear trajectories to better preserve vascular structure. Formally, the output at location $x_0$ is defined as:
\begin{equation}
y(x_0) = \sum_{i=1}^{K} w_i \cdot x\!\left(x_0 + p_i + \Delta p_i\right),
\end{equation}
where $K$ is the kernel size (number of sampling points), $w_i$ are convolution weights, $p_i$ are fixed grid locations, and $\Delta p_i$ are learnable offsets predicted from the input.

Compared with other dynamic convolutions such as the deformable convolution~\cite{dai2017deformable}, DSC models offset evolution along vessel-like paths, producing more stable alignment with elongated structures while remaining computationally tractable. Nevertheless, it introduces extra overhead. To balance accuracy and efficiency, we use DSC only once at the Stage~2 downsampling layer, based on three considerations. First, downsampling is where feature degradation is most severe, making it the critical point for structural preservation~\cite{ronneberger2015u,dai2017deformable,qi2023dynamic}. Second, placing DSC at higher resolutions substantially increases computation~\cite{dai2017deformable}. Third, after the transition to Vision Transformers (ViTs), the features are reorganized into patch tokens and local continuity becomes less pertinent; consequently, the DSC module offers limited additional benefit in that stage~\cite{dosovitskiy2020image,liu2021swin}. Accordingly, we apply a single DSC at the Stage~2 downsampling layer, right before passing the features into the ViT block. This design helps preserve vessel continuity at the stage where structural information is most likely to be lost, while at the same time avoiding the high computational cost of applying DSC at higher resolutions, and providing ViT with more coherent inputs for global modeling.


\subsection{Vision Transformer Stage}
In the ViT stage, we follow the MobileViT~\cite{mehta2021mobilevit} design, where convolution is first applied for downsampling and local feature aggregation, and the resulting features are then processed by lightweight Transformer blocks for global dependency modeling, as shown in Fig.~\ref{fig:arch}(c). Specifically, the input feature map $X \in \mathbb{R}^{H \times W \times C}$ is reduced to a compact representation $X' \in \mathbb{R}^{H' \times W' \times C'}$ via convolution, which is then projected into a sequence of tokens with positional encodings and fed into Transformer layers:
\begin{equation}
Z = \mathrm{Transformer}(X'),
\end{equation}
where each Transformer layer consists of Multi-Head Self-Attention (MHSA) and a Feed-Forward Network (FFN):
\begin{equation}
Z' = \mathrm{FFN}(\mathrm{MHSA}(Z)).
\end{equation}

Finally, the outputs of the convolution and Transformer branches are fused before being passed to the subsequent stage. Concretely, a local projection is first applied on $X$ and then added to $Z'$:
\begin{equation}
\hat{X} = \mathrm{Conv}_{\text{local}}(X),
\end{equation}
\begin{equation}
Y = \hat{X} + Z',
\end{equation}
where $Y$ denotes the fused feature map. This hybrid design enables convolution layers to capture fine local structures, while Transformer modules model long-range dependencies, together yielding representations that remain computationally compact while retaining strong descriptive capacity for retinal images.

\subsection{Self-supervised Pretraining on Retinal Images}
\paragraph{Self-supervised pretraining.}
To enhance generalization and stabilize optimization, we adopt masked image modeling for self-supervised pretraining. In particular, we use SimMIM~\cite{xie2022simmim}, which reconstructs randomly masked regions from the visible context without requiring architectural changes, making it naturally compatible with our hybrid CNN--ViT network.

Formally, let $X\!\in\!\mathbb{R}^{H\times W\times C}$ be an input image and $M\!\in\!\{0,1\}^{\tilde H\times \tilde W}$ a binary mask over patches (or pixels). Denote the index set of masked locations by $\mathcal{I}_M=\{\,i\mid M_i=1\,\}$. The model predicts the masked content $\hat X$, and the reconstruction loss is:
\begin{equation}
\mathcal{L}_{\text{SimMIM}}
=\frac{1}{|\mathcal{I}_M|}\sum_{i\in \mathcal{I}_M}\big\|\hat X_i - X_i\big\|_{1}.
\end{equation}
Here, $i$ indexes a masked patch (or pixel); $X_i$ and $\hat X_i$ are the ground-truth and predicted RGB vectors for that location; $|\mathcal{I}_M|$ is the number of masked locations; and $\|\cdot\|_{1}$ is the $L_1$ (MAE) norm. Only masked positions contribute to the loss; visible regions are ignored.

Through this strategy, the model learns retinal vascular priors from large-scale unlabeled images and acquires more transferable representations, thereby improving downstream retinal analysis performance.

\begin{figure}[h]
  \centering
\centerline{\includegraphics[width=0.35\textwidth]{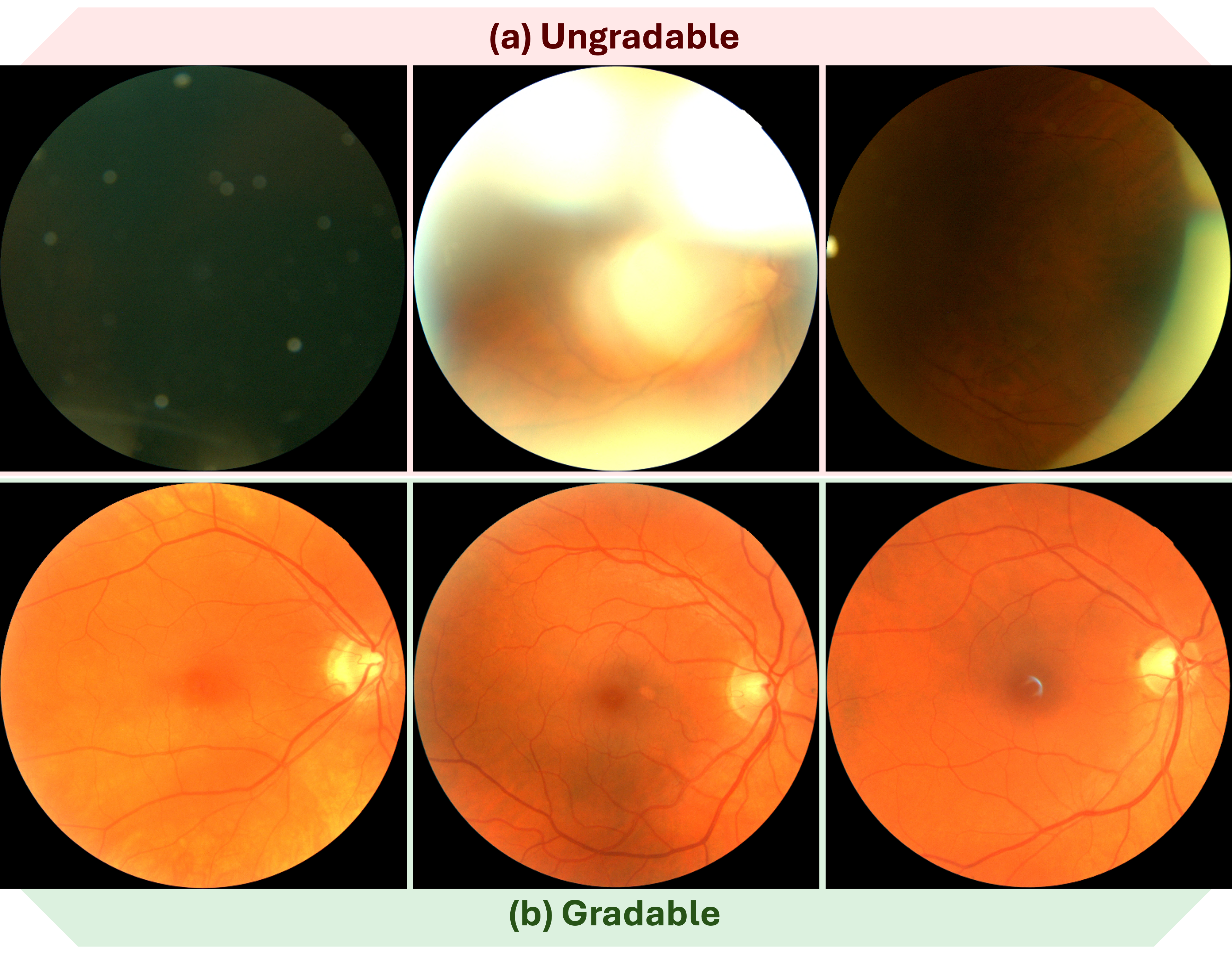}}
\caption{Examples of image quality in fundus images. 
The first row shows ungradable images excluded due to blur, poor illumination, or severe artifacts. 
The second row shows gradable images.}
\label{fig:quality_examples}
\end{figure}

\section{Experiments}
\subsection{Datasets}
We evaluate nnMobileNet++ across six retinal datasets that span multi-label classification, multi-class categorization, and challenging MICCAI benchmarks involving ultra-widefield imaging and cross-device generalization. 

\noindent \textbf{UK Biobank~\cite{bycroft2018uk}.}  
For self-supervised pretraining, we use the UK Biobank, a large-scale cohort comprising 179,127 retinal color fundus photographs (CFPs). After applying the AutoMorph~\cite{automorph} pipeline for quality control, 114,275 gradable CFPs were retained. This large dataset provides sufficient diversity to learn robust retinal priors.  

\noindent \textbf{Multi-Label Retinal Diseases (MuReD)~\cite{MuReD2022}.}  
MuReD contains 2,208 CFPs (1,746 training and 444 test images). Each image is annotated with a 20-dimensional multi-hot label vector, covering 19 retinal disease categories plus an ``other'' class. This design allows multiple diseases to co-occur in one image, reflecting realistic clinical settings.  

\noindent \textbf{Ocular Disease Intelligent Recognition (ODIR)~\cite{ODIR2019}.}  
The ODIR dataset includes 7,000 CFPs from around 3,500 patients, annotated with eight disease categories: diabetes, glaucoma, cataract, age-related macular degeneration, hypertension, pathological myopia, other abnormalities, and normal. We follow the official DGCAHMO order and treat this as a multi-class disease classification task.  

\noindent \textbf{MICCAI 2023 Myopic Maculopathy Analysis Challenge (MMAC)~\cite{MMAC2023}.}  
The MMAC dataset provides 1,143 training and 248 validation images for grading myopic maculopathy severity into five levels, from tessellated fundus (grade 1) to macular atrophy (grade 4). This progression reflects increasing structural damage and offers a clinically meaningful benchmark for evaluating automated methods.  

\noindent \textbf{MICCAI 2024 Ultra-Widefield Fundus Imaging for Diabetic Retinopathy (UWF4DR) Challenge~\cite{UWF4DR2024}.}  
UWF4DR consists of three tasks: image quality assessment, referable diabetic retinopathy (DR) identification, and diabetic macular edema (DME) detection. Unlike conventional CFPs (30°–50° FoV), ultra-widefield (UWF) images capture up to 200° of the retina, making peripheral signs visible but also introducing peripheral distortion, uneven illumination, and greater anatomical variability.  

\noindent \textbf{MICCAI 2025 Multi-Camera Robust Diagnosis of Fundus Diseases (MuCaRD)~\cite{MuCaRD2025}.}  
MuCaRD includes 400 images with diabetic retinopathy and 400 with glaucoma. Since only the training set has been released, we adopt a 5-fold cross-validation protocol. The multi-camera acquisition setting introduces substantial variability in image quality and device characteristics, providing a stringent test of model robustness.  

\begin{figure}[h]
  \centering
\centerline{\includegraphics[width=0.38\textwidth]{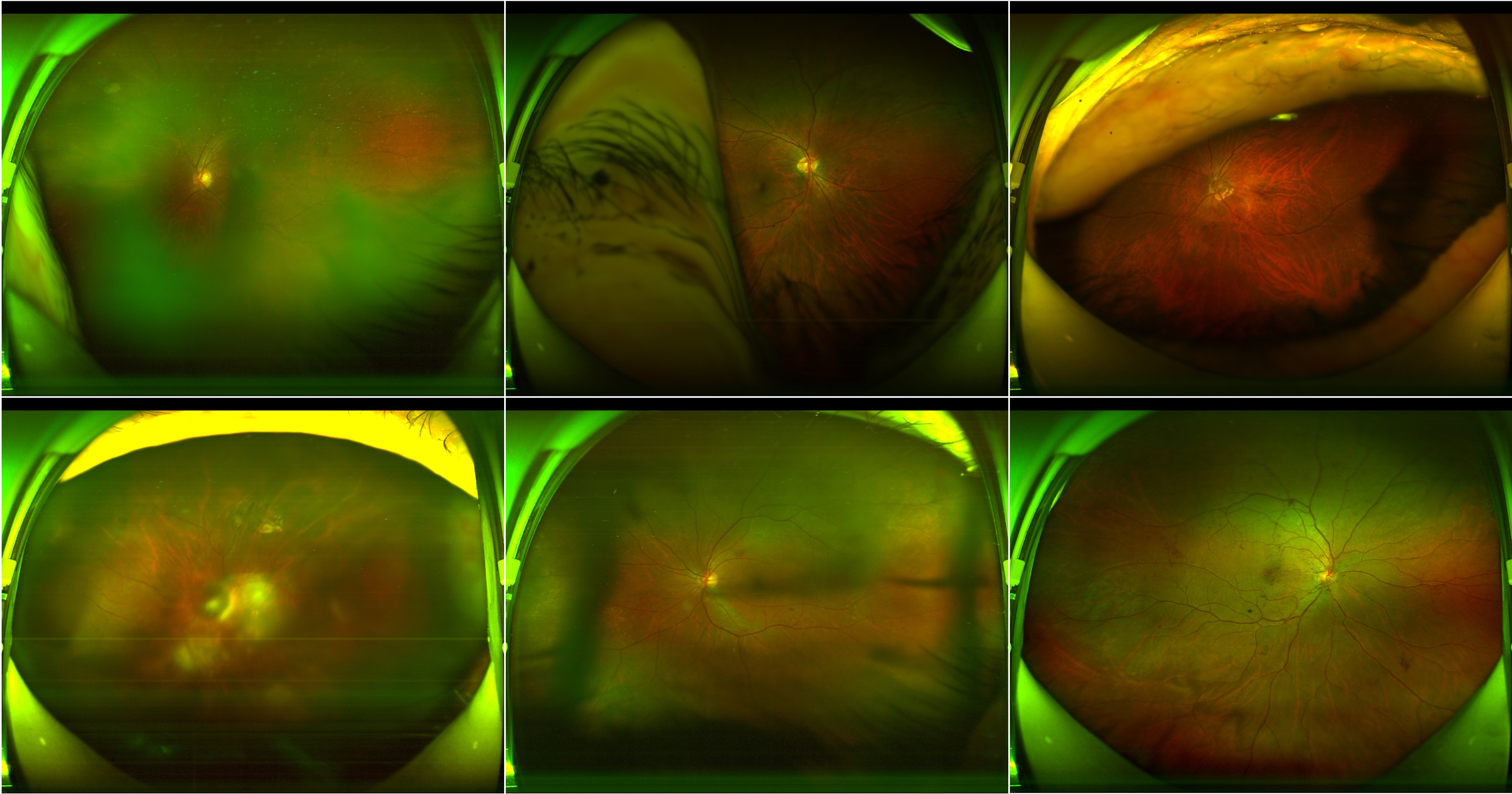}}
\caption{Examples of ultra-widefield (UWF) fundus images from the challenge dataset. The diversity of these images introduces challenges such as peripheral distortion, illumination non-uniformity, anatomical variability, and device/domain shifts.}
\label{fig:UWF}
\end{figure}

\begin{table*}[h]
\centering
\caption{\textbf{Multi-Label Retinal Diseases (MuReD) dataset.}
We report AUC, F1, and AUPRC; AUPRC is emphasized due to severe class imbalance. 
The righ columns summarize computational cost and model size—\emph{FLOPs (GMac)} and \emph{Params (M)}—to assess deployability~\cite{thop}. 
“Train from scratch” indicates models trained without external pretraining; \textit{“SSL” denotes self-supervised pretraining on retinal images for nnMobileNet++.}
}

\label{tab:mured_results}
\resizebox{0.85\textwidth}{!}{ 

\begin{tabular}{lcccccc}
\hline
\textbf{MIRel} & \textbf{AUC} & \textbf{F1} & \textbf{AUPRC} & \textbf{Train from scratch} & \textbf{FLOPs (GMac)} & \textbf{Params (M)} \\
\hline
ConvNeXt-Tiny            & 0.663 & 0.063 & 0.136 & Yes & 4.455 & 27.801 \\
EfficientNetV2-S    & 0.728 & 0.255 & 0.217 & Yes & 2.849 & 20.03 \\
Inception-V4        & 0.731 & 0.196 & 0.204 & Yes & 6.123 & 41.087 \\
LeViT               & 0.663 & 0.108 & 0.176 & Yes & 2.250 & 37.588 \\
nnMobileNet         & 0.690 & 0.091 & 0.158 & Yes & 0.428 & 3.522 \\
Swin Transformer-Tiny    & 0.714 & 0.218 & 0.244 & Yes & 4.371 & 27.500 \\
ViT-Tiny            & 0.684 & 0.073 & 0.167 & Yes & 1.075 & 5.478 \\
ViT-Base            & 0.682 & 0.053 & 0.173 & Yes & 16.848 & 85.612 \\
MobileViT           & 0.718 & 0.123 & 0.188 & Yes & 1.420 & 4.929 \\
MobileNetV2         & 0.690 & 0.091 & 0.158 & Yes & 0.3 & 2.196 \\
\hline
\textbf{nnMobileNet++*}          & \textbf{0.805} & \textbf{0.334} & \textbf{0.335} & \textbf{Yes}  & \textbf{2.538} & \textbf{2.1} \\
\textbf{nnMobileNet++} & \textbf{0.842} & \textbf{0.445} & \textbf{0.398} & \textbf{SSL}  & \textbf{2.538} & \textbf{2.1} \\

\hline
\end{tabular}
}
\end{table*}

\begin{table}[h]
\centering
\caption{\textbf{ODIR Dataset.} Results of multi-disease classification. We report AUC, F1, and Accuracy for baseline models and nnMobileNet++. nnMobileNet++* indicates training from scratch.}
\label{tab:odir_results}
\resizebox{0.4\textwidth}{!}{ 

\begin{tabular}{lccc}
\toprule
\textbf{Model} & \textbf{AUC} & \textbf{F1} & \textbf{Accuracy} \\
\midrule
ConvNeXt         & 0.854 & 0.270 & 0.441 \\
nnMobileNet      & 0.856 & 0.270 & 0.441 \\
EfficientNetV2-S & 0.855 & 0.270 & 0.441 \\
Inception-V4     & 0.855 & 0.270 & 0.441 \\
LeViT            & 0.859 & 0.272 & 0.442 \\
MobileNetV2      & 0.869 & 0.304 & 0.449 \\
MobileViT        & 0.868 & 0.288 & 0.448 \\
Swin-Transformer & 0.856 & 0.277 & 0.442 \\
\midrule
\textbf{nnMobileNet++*} & \textbf{0.873} & \textbf{0.295} & \textbf{0.445} \\
\textbf{nnMobileNet++} & \textbf{0.906} & \textbf{0.494} & \textbf{0.536} \\
\bottomrule
\end{tabular}
}
\end{table}

\begin{table*}[h]
\centering
\caption{\textbf{MICCAI 2024 UWF4DR dataset.} Results on ultra-widefield (UWF) fundus images, which present greater challenges due to peripheral distortion, illumination variability, and anatomical diversity. We evaluate three tasks: image quality assessment, diabetic retinopathy (DR) classification, and diabetic macular edema (DME) classification. Across all tasks, nnMobileNet++ achieves the best performance, consistently surpassing CNN and ViT baselines in terms of AUC, Accuracy, and F1.}
\label{tab:uwf4dr_results}
\resizebox{0.78\textwidth}{!}{ 
\begin{tabular}{l|ccc|ccc|ccc}
\toprule
\multirow{2}{*}{\textbf{Model}} & \multicolumn{3}{c|}{\textbf{Image Quality Assessment}} & \multicolumn{3}{c|}{\textbf{DR Classification}} & \multicolumn{3}{c}{\textbf{DME Classification}} \\
\cmidrule(lr){2-4}\cmidrule(lr){5-7}\cmidrule(lr){8-10}
 & \textbf{AUC} & \textbf{ACC} & \textbf{F1} & \textbf{AUC} & \textbf{ACC} & \textbf{F1} & \textbf{AUC} & \textbf{ACC} & \textbf{F1} \\
\midrule
ConvNeXt          & 0.846 & 0.787 & 0.787 & 0.762 & 0.680 & 0.680 & 0.775 & 0.644 & 0.644 \\
EfficientNetV2\text{-}S & 0.826 & 0.754 & 0.754 & 0.665 & 0.640 & 0.627 & 0.773 & 0.711 & 0.711 \\
Inception\text{-}V4     & 0.840 & 0.738 & 0.738 & 0.770 & 0.740 & 0.740 & 0.807 & 0.756 & 0.751 \\
LeViT              & 0.848 & 0.738 & 0.705 & 0.638 & 0.580 & 0.426 & 0.621 & 0.533 & 0.371 \\
nnMobileNet        & 0.834 & 0.754 & 0.754 & 0.678 & 0.600 & 0.600 & 0.829 & 0.711 & 0.703 \\
MobileNetV2        & 0.841 & 0.754 & 0.727 & 0.665 & 0.640 & 0.610 & 0.696 & 0.689 & 0.689 \\
MobileViT          & 0.822 & 0.770 & 0.748 & 0.682 & 0.640 & 0.599 & 0.647 & 0.644 & 0.641 \\
Swin\text{-}Transformer & 0.703 & 0.607 & 0.607 & 0.716 & 0.580 & 0.580 & 0.666 & 0.600 & 0.600 \\
ViT\text{-}Base         & 0.772 & 0.738 & 0.719 & 0.645 & 0.580 & 0.426 & 0.585 & 0.533 & 0.371 \\
ViT\text{-}Tiny         & 0.824 & 0.787 & 0.763 & 0.623 & 0.580 & 0.426 & 0.587 & 0.600 & 0.589 \\
\midrule
\textbf{nnMobileNet++} & \textbf{0.854} & \textbf{0.803} & \textbf{0.799} & \textbf{0.882} & \textbf{0.840} & \textbf{0.841} & \textbf{0.893} & \textbf{0.867} & \textbf{0.867} \\
\bottomrule
\end{tabular}
}
\end{table*}

\subsection{Implementation Details}
 In both pretraining and classification tasks, input images were uniformly resized to 224×224 and trained with AdamW~\cite{wang2025adawm} optimizer (initial learning rate 1e-3, cosine decay schedule, and batch size 32), with standard normalization. Pretraining applied 60\% random patch masking (patch size 32) and was run for 800 epochs with weight decay 0.05 and 20 warmup epochs, using automatic mixed precision (AMP). For classification, all CNN and ViT baselines were trained from scratch using the timm library~\cite{rw2019timm}, while nnMobileNet++ was evaluated both from scratch and with a pretrained ReXNet backbone, trained for 300 epochs with weight decay 5e-4. All experiments were implemented in PyTorch and executed on NVIDIA A100 GPUs(80GB).

\begin{table}[h]
\centering
\caption{\textbf{MICCAI 2023 MMAC dataset.} 
We report AUC, F1, and Accuracy for various baseline models and nnMobileNet++.}
\label{tab:mmac}
\begin{tabular}{lccc}
\toprule
\textbf{Model} & \textbf{AUC} & \textbf{F1} & \textbf{Accuracy} \\
\midrule
ConvNeXt       & 0.572 & 0.513 & 0.513 \\
EfficientNetV2-S & 0.878 & 0.565 & 0.593 \\
Inception-V4   & 0.783 & 0.411 & 0.448 \\
MobileNetV2    & 0.855 & 0.554 & 0.585 \\
MobileViT      & 0.878 & 0.588 & 0.621 \\
Swin-Transformer & 0.742 & 0.219 & 0.351 \\
nnMobileNet    & 0.919 & 0.645 & 0.669 \\
\midrule
\textbf{nnMobileNet++} & \textbf{0.946} & \textbf{0.746} & \textbf{0.758} \\
\bottomrule
\end{tabular}
\end{table}

\begin{table}[h]
\centering
\caption{Ablation study conducted on the MICCAI MMAC 2023 dataset. 
$B$: nnMobileNet backbone; 
$V$: replacing the IRLB block with a ViT module; 
$S$: adding Dynamic Snake Conv (DSC) in place of the IRLB convolution; 
$P$: applying pretraining on retinal images. 
Performance is reported in terms of AUC, F1 score, FLOPs (GMac), and Params (M).}
\resizebox{0.4\textwidth}{!}{ 
\begin{tabular}{c@{\hskip 5pt}|c@{\hskip 5pt}c@{\hskip 5pt}c@{\hskip 5pt}c@{\hskip 5pt}|c@{\hskip 6pt}c@{\hskip 6pt}c@{\hskip 6pt}c@{\hskip 6pt}}
\hline
\textbf{Index} & \textbf{$B$} & \textbf{$V$} & \textbf{$S$} & \textbf{$P$} & \textbf{AUC} & \textbf{F1} & \textbf{FLOPs} & \textbf{Params}\\
\hline
1 & $\checkmark$ & $\times$      & $\times$      & $\times$      & 0.919 & 0.645 & 0.428 & 3.522 \\
2 & $\checkmark$ & $\checkmark$  & $\times$      & $\times$      & 0.928 & 0.691 & 0.603 & 1.933 \\
3 & $\checkmark$ & $\checkmark$  & $\checkmark$  & $\times$      & 0.935 & 0.730 & 2.538 & 2.100 \\
4 & $\checkmark$ & $\checkmark$  & $\checkmark$  & $\checkmark$  & 0.946 & 0.746 & 2.538 & 2.100 \\
\hline
\end{tabular}%
}
\label{tab:ablation}
\end{table}

\begin{table}
\centering
\caption{\textbf{MICCAI 2025 MuCaRD dataset.} This dataset is characterized by its multi-device acquisition setting. We report AUC, AUPRC, and F1 as evaluation metrics for baseline models and nnMobileNet++.}
\resizebox{0.4\textwidth}{!}{ 

\begin{tabular}{lrrr}
\toprule
\textbf{Model}      & \textbf{AUC}   & \textbf{AUPRC} & \textbf{F1} \\
\midrule
ConvNeXt   & 0.572 & 0.575 & 0.513 \\
EfficientNetV2-S        & 0.629 & 0.630 & 0.513 \\
Inception-V4  & 0.645 & 0.625 & 0.513 \\
Swin-Transformer       & 0.598 & 0.601 & 0.513 \\
LeViT       & 0.588 & 0.595 & 0.513 \\
MobileNetV2      & 0.565 & 0.591 & 0.538 \\
MobileViT      & 0.609 & 0.595 & 0.513 \\
nnMobileNet    & 0.659 & 0.636 & 0.513 \\
ViT-Tiny    & 0.563 & 0.561 & 0.513 \\
ViT-Base   & 0.580 & 0.588 & 0.525 \\
\midrule
\textbf{nnMobileNet++} & \textbf{0.739} & \textbf{0.725} & \textbf{0.669} \\
\bottomrule
\label{tab:mu_challange}
\end{tabular}
}
\end{table}

\begin{figure*}[h]
  \centering
  \centerline{\includegraphics[width=1\textwidth]{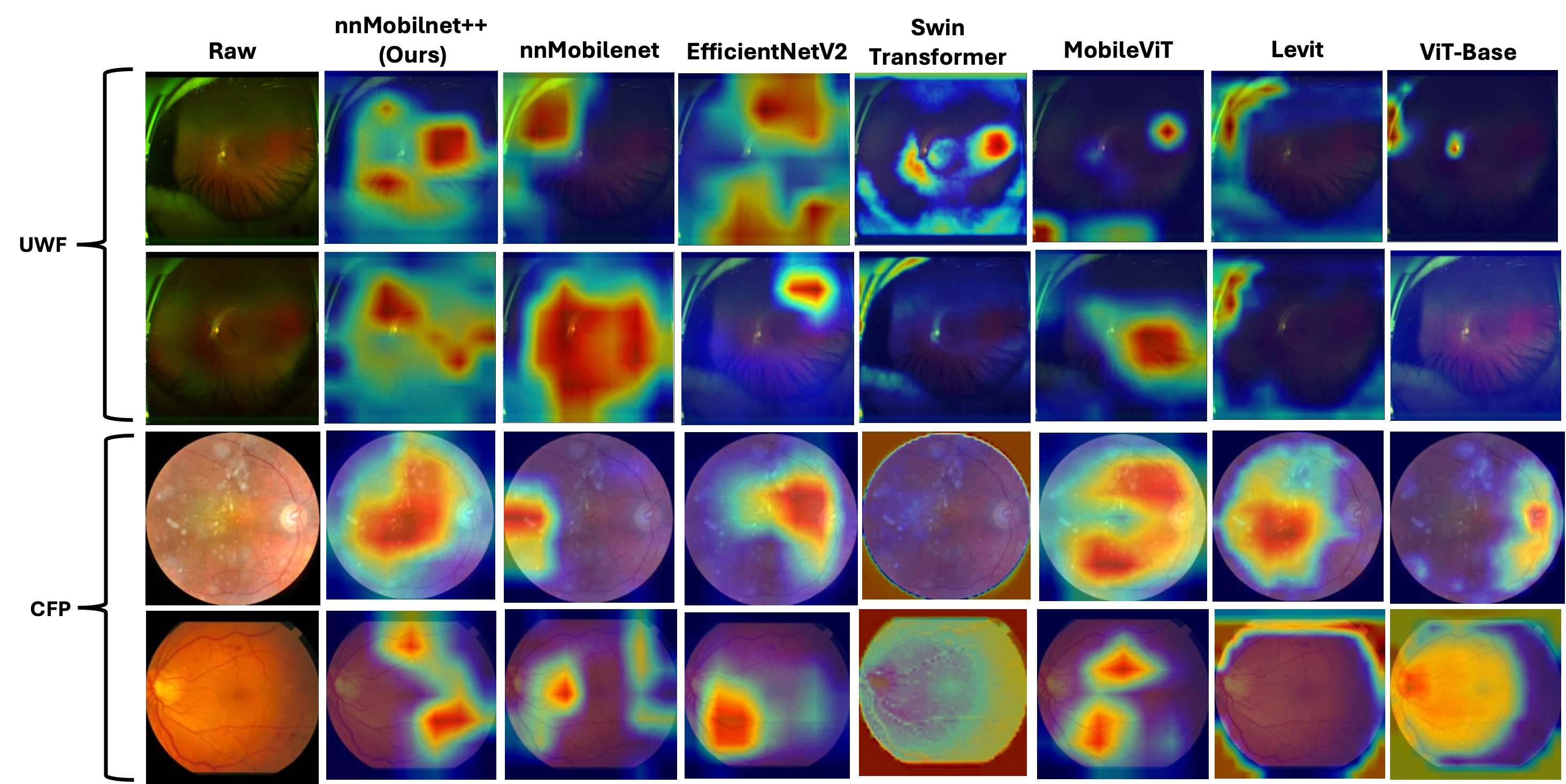}}
\caption{
Class activation maps (CAMs)~\cite{selvaraju2017grad} generated by different networks. The first two rows show Ultra-Widefield Fundus Imaging (UWF) samples from the UWF4DR dataset, and the last two rows show color fundus photography (CFP) samples from the MURED dataset. Compared to existing baselines (nnMobilenet~\cite{zhu2024nnmobilenet}, EfficientNetV2~\cite{tan2019efficientnet}, Swin Transformer~\cite{liu2021swin}, MobileViT~\cite{mehta2021mobilevit}, LeViT~\cite{graham2021levit}, and ViT-Base~\cite{dosovitskiy2020image}), our proposed nnMobileNet++ achieves consistently more accurate and focused attention on lesion regions across both UWF and CFP modalities.
}
\label{fig:cam}
\end{figure*}

\subsection{Main Experiments}
\noindent\textbf{Multi-Label Classification Results.}  
We evaluate a diverse set of CNN and ViT baselines alongside our proposed nnMobileNet++ on the MuRED dataset under a unified protocol. To ensure a fair comparison, all models are trained from scratch to isolate architectural capacity; we additionally perform self-supervised pretraining on retinal images to assess further gains on nnMobileNet++. As summarized in Table.~\ref{tab:mured_results}, the scratch version of nnMobileNet++ obtains AUC 0.805 / F1 0.334 / AUPRC 0.335, outperforming the strongest baselines (Inception-V4 for AUC 0.731, EfficientNetV2-S for F1 0.255, and Swin-Tiny for AUPRC 0.244) by +0.074 AUC / +0.079 F1 / +0.091 AUPRC. With self-supervised pretraining, nnMobileNet++ further improves to AUC 0.842 / F1 0.445 / AUPRC 0.398. In terms of efficiency, nnMobileNet++ requires only 2.5 GMacs FLOPs and 2.1M parameters, making it lightweight and suitable for deployment on resource-constrained devices; its FLOPs are substantially lower than mainstream ViT backbones such as Swin-Tiny (4.4 GMacs and 27.5M Parameters). These results indicate that the proposed architecture provides strong performance while remaining computationally compact, and that pretraining offers additional benefit in data-limited medical imaging scenarios.

\noindent\textbf{Multi-Class Classification Results.}
On the ODIR dataset for multi-disease, multi-class retinal classification, baseline CNN and ViT models achieve AUC values around 0.85–0.87 but consistently suffer from very low F1 scores ($\leq0.304$), highlighting the difficulty of multi-class prediction under class imbalance. As shown in Table.~\ref{tab:odir_results}Our nnMobileNet++ trained from scratch achieves an AUC of 0.873, F1 of 0.295, and Accuracy of 0.445, already matching or slightly surpassing the best baselines. With self-supervised pretraining and fine-tuning, nnMobileNet++ further improves to an AUC of 0.906, F1 of 0.494, and Accuracy of 0.536. This represents improvements of +0.033 AUC, +0.199 F1, and +0.091 Accuracy over its scratch counterpart, while substantially outperforming all baselines. These results indicate that nnMobileNet++ is highly effective for challenging multi-class, imbalanced retinal classification, and that self-supervised pretraining provides considerable additional benefit.

\noindent\textbf{MICCAI Challenge Datasets Results.}
We further evaluate nnMobileNet++ on three recent MICCAI challenges, which provide clinically realistic and increasingly demanding benchmarks beyond standard fundus datasets. On \textbf{MMAC 2023}, which targets the grading of myopic maculopathy, disease progression is reflected in subtle morphological changes of the posterior pole that are often difficult for lightweight models to capture. As shown in Table~\ref{tab:mmac}, nnMobileNet++ achieves the best results (AUC 0.946 / F1 0.746 / Accuracy 0.758), demonstrating its capacity to capture fine-grained pathological variations more effectively than conventional CNN and ViT baselines. On \textbf{UWF4DR 2024}, which involves ultra-widefield (UWF) fundus images covering up to 200° of the retina~\ref{fig:UWF}, peripheral distortion, uneven illumination, and anatomical variability present major challenges. As reported in Table~\ref{tab:uwf4dr_results}, nnMobileNet++ consistently exceeds baselines in three tasks: image quality assessment, classification of diabetic retinopathy (DR), and classification of diabetic macular edema (DME). For example, it achieves an AUC of 0.893 and F1 of 0.867 on DME classification, confirming strong robustness to UWF-specific artifacts. On \textbf{MuCaRD 2025}, designed to assess robustness across multiple imaging devices, substantial domain shifts in resolution, color, and illumination substantially degrade baseline performance. As shown in Table~\ref{tab:mu_challange}, nnMobileNet++ obtains the highest scores (AUC 0.739 / AUPRC 0.725 / F1 0.669), exceeding nnMobileNet by +0.080 AUC and +0.156 F1. These gains highlight its improved generalization under device heterogeneity, which is essential for real-world deployment. In summary, nnMobileNet++ achieves strong predictive performance while maintaining a lightweight design. It also shows robustness to disease heterogeneity, anatomical distortions, and device variability. This balance of accuracy and efficiency makes it well-suited for retinal image analysis in real-world clinical settings, especially under resource constraints.

\subsection{Ablation Studies}
We conduct ablation experiments on the \textbf{MICCAI MMAC 2023} dataset to assess the impact of each component in nnMobileNet++ (Table.~\ref{tab:ablation}). Starting from the nnMobileNet backbone (AUC 0.919 / F1 0.645), replacing the IRLB block with a ViT module improves performance (AUC 0.928 / F1 0.691), confirming the value of global context modeling. Incorporating DSC further raises accuracy (AUC 0.935 / F1 0.730), demonstrating its effectiveness in maintaining vessel continuity. With pretraining, the model achieves the best results (AUC 0.946 / F1 0.746).

Overall, nnMobileNet++ enhances performance while reducing parameter count (3.5M $\rightarrow$ 2.1M) and keeping FLOPs modest ($\leq$2.5 GMacs). Importantly, the variant without DSC also performs competitively (AUC 0.928 / F1 0.691) with only 0.6 GMacs, showing that the framework remains highly efficient under tight computational budgets. We further validate this on the \textbf{MuReD} dataset (Fig.~\ref{fig:flops}), where the variant without DSC but with pretraining still achieves strong AUC results. These findings indicate that nnMobileNet++ balances accuracy and efficiency, making it suitable for deployment across diverse environments.

\subsection{Grad-CAM Visualization}
Fig.~\ref{fig:cam} presents Grad-CAM~\cite{selvaraju2017grad} visualizations comparing nnMobileNet++ with several baselines. On the challenging UWF images (top two rows), nnMobileNet++ is still able to focus on lesion regions, whereas other models often fail to localize meaningful areas. On CFP images (bottom two rows), our model demonstrates superior precision, especially when lesions are small or visually subtle. For instance, in the last row, two faint DR lesions—likely corresponding to retinal hemorrhages or microaneurysms, which are difficult to detect due to their small size and color similarity to the surrounding retina—are correctly highlighted only by nnMobileNet++. Such visualizations are of great clinical importance, as they not only confirm the correctness of model predictions but also provide interpretability for subtle and clinically relevant lesions.

\section{Conclusion}
We presented nnMobileNet++, a lightweight hybrid CNN–ViT framework enhanced with Dynamic Snake Convolution for retinal image analysis. Across MuReD, ODIR, and three MICCAI challenges, it consistently surpasses CNN and ViT baselines while remaining highly efficient. Grad-CAM visualizations confirm reliable localization of subtle lesions, supporting interpretability. These advances suggest nnMobileNet++ can provide accurate and scalable tools for retinal screening. Beyond technical performance, its efficiency and robustness make it particularly suited for integration into clinical research pipelines, where it may facilitate the discovery of imaging biomarkers and support precision medicine in ophthalmology and systemic disease.

{
    \small
    \bibliographystyle{ieeenat_fullname}
    \bibliography{main}
}

\end{document}